\documentclass[conference]{IEEEtran}
\IEEEoverridecommandlockouts
\usepackage{cite}
\usepackage{amsmath,amssymb,amsfonts}

\newtheorem{definition}{\bfseries Definition}
\newtheorem{problem}{\bfseries Problem}

\usepackage{algorithm}
\usepackage{algorithmic}
\usepackage{graphicx}
\usepackage{textcomp}
\usepackage{xcolor}
\usepackage{booktabs}
\usepackage{subfigure}

\newcommand{\eg}{{\em e.g.,~}} 
\newcommand{\ie}{{\em i.e.,~}}

\def\BibTeX{{\rm B\kern-.05em{\sc i\kern-.025em b}\kern-.08em
    T\kern-.1667em\lower.7ex\hbox{E}\kern-.125emX}}
\begin{document}

\title{Federated Learning Incentive Mechanism under Buyers’ Auction Market}

\author{Jiaxi Yang$^{1,\dag}$\thanks{$^\dag$Jiaxi Yang and Zihao Guo contributed equally to this work.}, Zihao Guo$^{2, \dag}$, Sheng Cao$^1$, Cuifang Zhao$^{3}$, Li-Chuan Tsai$^{4,*}$ \thanks{$^*$Correspondence to: Li-Chuan Tsai $<$1201500112@jxufe.edu.cn$>$}  \\
$^1$University of Electronic Science and Technology of China \\
$^2$Beihang University, $^3$National Taiwan University\\ 
$^4$Jiangxi University of Finance and Economics \\

}

\maketitle

\begin{abstract}

Auction-based Federated Learning (AFL) enables open collaboration among self-interested data consumers and data owners. Existing AFL approaches are commonly under the assumption of sellers' market in that the service clients as sellers are treated as scarce resources so that the aggregation servers as buyers need to compete the bids. Yet, as the technology progresses, an increasing number of qualified clients are now capable of performing federated learning tasks, leading to shift from sellers' market to a buyers' market. In this paper, we shift the angle by adapting the procurement auction framework, aiming to explain the pricing behavior under buyers' market. Our modeling starts with basic setting under complete information, then move further to the scenario where sellers' information are not fully observable. In order to select clients with high reliability and data quality, and to prevent from external attacks, we utilize a blockchain-based reputation mechanism. The experimental results validate the effectiveness of our approach.

\end{abstract}

% \begin{IEEEkeywords}
% component, formatting, style, styling, insert
% \end{IEEEkeywords}

\section{Introduction}
Due to the costs (e.g., the risk of privacy leakage and consumption of computation resources) for clients to participate in federated learning (FL) tasks, incentive mechanism design for FL has received significant research interest~\cite{zeng2021comprehensive}. As one of the efficient methods to address this issue, auction-based FL (AFL) is a promising approach that has received a lot of attention.
As a buyer, the aggregation server recruits several clients to contribute their local data and computation resources to help complete the FL tasks.
% each aggregation server faces several clients and the aggregation server takes on the role of buyer and provides incentives to recruit clients to 
According to the objective of optimization, existing studies can be divided into three categories~\cite{tang2023utility}: \textit{(1) maximize profit of buyers:} the aim of the aggregation server here is to maximize its own utility by efficiently recruiting high-quality clients and ensuring the model training process converges quickly to obtain an effective global model; \textit{(2) maximize profit of sellers:} the clients determine how much of their local data and computation resources they are willing to contribute to the FL tasks. Each client bids to maximize their own expected profit from participating in; \textit{(3) maximize profit of FL community:} the problem is formulated to server-client matching and pricing and optimize the utility of the whole FL community.

\begin{figure}[]
  \centering
  \begin{subfigure}
  \centering{\includegraphics[scale=0.6]{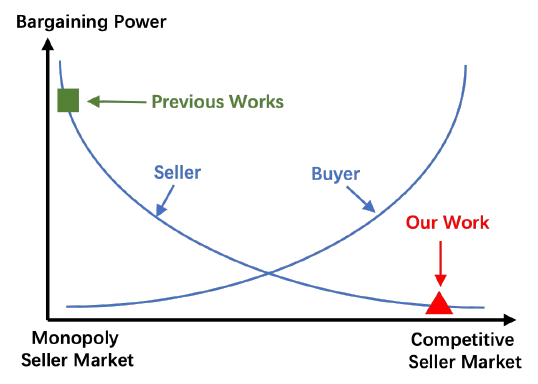}}
  \label{fig:subfig1}
  \end{subfigure}
  
  \caption{The market transition from monopoly seller market to competitive seller market.}
  \label{market}
\end{figure}

Although aforementioned methods have their own merit considerations, they all suffered a constrained assumption: \textit{the service providers (clients) are considered scarce resources, requiring the aggregation servers (buyers) to compete for recruiting them}.
In other words, the clients constitute the seller market, in which bargaining power is positively related to their scarcity. Above situation, however, will be changed when the number of qualified clients increase to certain extent. The pricing behavior will be different and the bargaining power may shift as the clients market become more and more competitive. Fig.~\ref{market} illustrates this phenomenon. Motivated by this observation, we attempt to reconsider the pricing behavior in buyers' market and take a different perspective by adopting the procurement auction.

% In other words, no single client can affect the market by changing its bidding behavior and clients only decide whether or not to accept the FL tasks in buyers' market. The bargaining power of clients in both sellers' market and buyers' market is represented in Fig.~\ref{market} that the bargaining power of clients tends to decrease when transitioning from a seller's market to a buyer's market. 

% \begin{figure}[t]
% 	\centering{\includegraphics[scale=0.4]{figures/market.png}}
%     \caption{}
% 	\label{market}
% \end{figure}

Considering information asymmetry between the aggregation server and clients, we compare the performance of our approach under both complete and incomplete information scenarios. And under the incomplete information setting, some private information (\eg efficiency) of clients is not completely observable, which lead to the allocation inefficiency for the aggregation server compared to complete information scenario. To alleviate this issue, the aggregation server needs to pay information rent for allocation efficiency increase. We try to explore this and discuss the trade-off problem that the aggregation server faces: \textit{sharing more information improves allocation efficiency but also leads to higher information rents}. 

To determine the winners selection and protect from potential security threats (\eg poisoning attack), the aggregation server need to select top-$k$ clients with high reputation. We further propose a blockchain based reputation mechanism to enhance the trustworthiness for the reputation record storage. The main contributions of our work are presented as follows:
\begin{itemize}
    \item To the best of our knowledge, we pioneer to explore AFL in buyers' market. 
    \item Due to the information asymmetry concern, we separately discuss the trade-off problem of information rent by clients under the both complete and incomplete information setting.
    \item We design a reputation mechanism to select candidate clients for the aggregation server. To make it more trustworthy, we use blockchain technology for reputation management.
    \item We perform extensive experiments comparing our approach to several baseline methods, and the results demonstrate the effectiveness of our proposed approach.
\end{itemize}

\section{Related Work}
Existing auction-based incentive mechanisms in Federated Learning (FL) can be categorized based on their optimization targets. The first two categories focus on maximizing the utility of clients and the FL community. Considering the competitive and cooperative relationship among clients, multi-agent reinforcement learning is applied to the auction process and achieve the maximum profit of clients.
Methods falling into another category employ various auction approaches, such as greedy-based auction and double auction, to determine the winners and maximize social welfare~\cite{zavodovski2019decloud, le2021incentive}. Others also aim to minimize social cost through procurement auctions in Non-IID settings of FL~\cite{zhou2021truthful}. Our work is is orthogonal to these works and addresses a different aspect of the problem.

The category addressed in this paper similarly aims to maximize the profits obtained by the aggregation server. Existing studies use procurement auction to tackle this, which involves one buyer (the aggregation server) and multiple sellers (clients) to maximize the utility of the aggregation server~\cite{zeng2020fmore, deng2021fair, zhang2021incentive}. Additionally, techniques such as reinforcement learning and graph neural networks are combined with procurement auctions to address this issue~\cite{jiao2020toward, zhang2021incentive}. Previous works assume a sellers' market where clients have some bargaining power and can adjust their compensation through their actions. However, this assumption becomes impractical with the increasing number of potential clients. As competitive degree rises, the market gradually shifts to a buyers' market.

% Existing auction-based incentive mechanisms in FL could be roughly grouped according to their optimization target. The first categories formulate the problem into maximize the utility of clients and obtain global optimization in the FL community. For auction-based incentive mechanism to optimize the utility of FL ecosystem, several kinds of auction approaches (\ie greedy-based auction, double auction) are leveraged to determine the winners and maximize the social welfare~\cite{zavodovski2019decloud, le2021incentive}. Others try to minimize the social cost based on procurement auction in Non-IID settings of FL~\cite{zhou2021truthful}. Our work is orthogonal to these works. 

% To achieve desirable objectives of the aggregation server, several studies leverage procurement auction, in which there is one buyer (the aggregation server) and multiple clients (sellers), for utility maximization of the aggregation server~\cite{zeng2020fmore, deng2021fair, zhang2021incentive}. Besides, various techniques including reinforcement learning, graph neural networks etc. are combined with reverse auction to tackle this problem~\cite{jiao2020toward, zhang2021incentive}. In these works, they all assume that the market is sellers' market in which the clients have bargaining power to some extent and are able to change its compensation by their actions. However, with the increasing number of available data owners which are eligible to be clients, the assumption of previous works is impractical. With the competitive pressure become much higher, the market gradually change to buyers' market.  

\section{System Model}
% In this section, we provide an outline of the overall workflow and give a detailed definition of the problem.

\begin{figure}[ht]
	\centering{\includegraphics[scale=0.45]{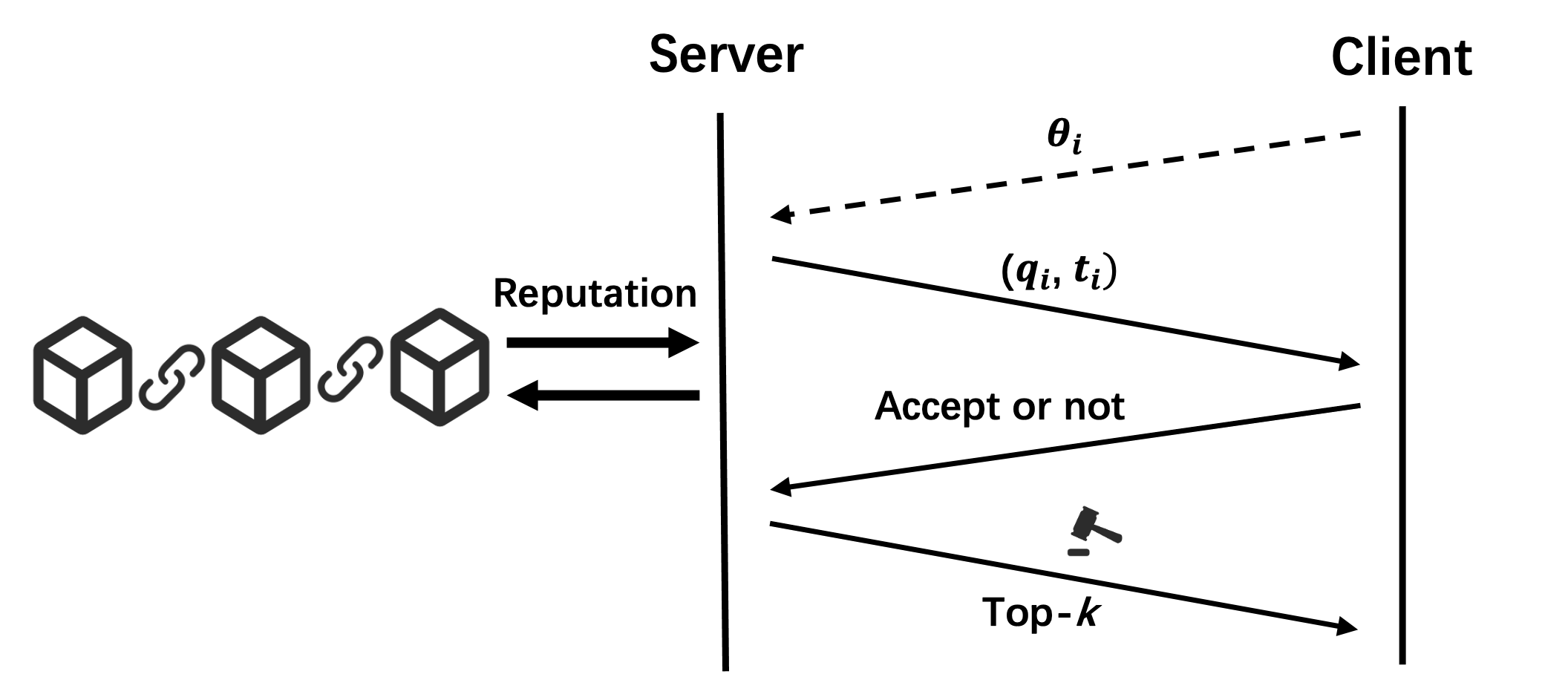}}
    \caption{Pipeline of our approach}
	\label{workflow}
\end{figure}

In FL ecosystem, there are typically two main parties: an aggregation server and multiple clients. 
The training process involves updating local models using clients' private data and computation resource, and the primary responsibility of the aggregation server is to recruit clients and coordinate them for model training in a decentralized manner. Therefore, the aggregation server plays a role as buyer, and pays for the work of clients which are regarded as sellers. The workflow of an auction is shown in Fig.~\ref{workflow}. The scenario involves a server and $N$ clients.
\textit{Firstly}, the clients have choice to reveal their private information (\eg efficiency $\theta_{i}$) to the aggregation server (Sec.~\ref{sec:complete} and Sec.~\ref{sec:incomplete}). The efficiency in implementing the project is $\theta\in[0,1]$, so the clients are regarded as more efficient when their efficiency parameter $\theta_{i}$ increases. \textit{Secondly}, the aggregation server chooses the output-transfer pair, $(q_{i}, R_{i})$, for each client $i$ to maximize its profit. The notation $q_{i}$ denotes the expected output of the client $i$ (\eg the improvement of test accuracy) \textit{Thirdly}, the client $i$ select whether to participate in, subject to the their rationality condition $U_{i}(q_{i}(\theta), \theta_{i}) \geq 0$ for all $\theta_{i} \in [0,1]$. \textit{Fourthly}, by providing historical reputation record from blockchain, the aggregation server selects the top-$k$ reputation clients in the auction.

\section{The Approach Under the Complete Information}
\label{sec:complete}
\subsection{Problem Formulation}

Consider a scenario where a server invites $N$ clients to participate in a computing contract. Initially, we assume that all clients are willing to disclose their private efficiency level, denoted as $\theta_{i}$, to the server. However, in the subsequent section~\ref{sec:incomplete}, we relax this assumption and allow clients $i$ to keep  their efficiency level $\theta_{i}$ private and unobservable to the aggregation server. The cost for client $i$ to implement the contract is denoted as $C_{i}(q_{i}, \theta_{i})$, which is increasing, convex in $q_{i}$, and also decreasing, convex in the efficiency level $\theta_{i}$ of client $i$. Specifically, we assume that the cost function of each client is:

\begin{equation}
    C_{i}(q_{i},\theta_{i}) = \frac{q_{i}^{2}}{1+\delta \cdot \theta_{i}}.
\end{equation}

The contract $C_{i}(q_{i}, \theta_{i})$ must satisfy the condition $\frac{\partial^{2}C_{i}(q_{i}, \theta_{i})}{\partial q_{i} \partial \theta_{i}} \leq 0$. The negative sign of the cross-partial derivative $\frac{\partial^{2}C_{i}(q_{i}, \theta_{i})}{\partial q_{i} \partial \theta_{i}}$ implies that as the efficiency of client $i$ increases, its marginal cost of computing decreases.
In other words, the value of $\frac{\partial^{2}C_{i}(q_{i}, \theta_{i})}{\partial q_{i} \partial \theta_{i}}$ decreases as $\theta_{i}$ increases. As a result, each client has a quasi-linear utility function, represented as follows:

\begin{equation}
    U_{client}(q_{i}, \theta_{i}) = R(q_{i}) - C_{i}(q_{i}, \theta_{i}),
\label{utilityOfClient}
\end{equation}

where $R(q_{i})$ represents the transfer (reward) that client $i$ receives from the aggregation server. $q_{i}$ denoted in equation (\ref{contribution}) is the contribution of client $i$ for model training, and we define it as the discrepancy between the test accuracy of model $\mathcal{M}$ before and after the local training.
For simplicity, assume that clients earn a zero reservation utility if they choose to not participate in the auction. Then the server’s utility function from the client $i$ is in the equation (\ref{complete:serverUtility}).

\begin{equation}
    q_{i} =  Acc(\mathcal{M}_{local}) - Acc(\mathcal{M}_{global})
    \label{contribution}
\end{equation}

\begin{equation}
    U_{server} = V(q_{i}) -  R(q_{i}), 
    \label{complete:serverUtility}
\end{equation}

where $V(q_{i})$ denotes the value that the server assigns to $q_{i}$, which can be denoted as $V(q_{i}) = \lambda \cdot q_{i}$.
The set of clients that participate in the FL task is denoted as $S=\{s_{1}, s_{2}, ..., s_{k}\}$. Thus the final utility function of the aggregation server is:

\begin{equation}
    U_{server} = \sum_{i=1}^{k} (V(q_{i}) - R(q_{i})).
    \label{}
\end{equation}

Under the complete information, the optimization problem of the aggregation server in buyers' market is formally given below.

\begin{problem}[Maximize server's utility function under the complete information]
\label{problem:MaxServerComplete}
	\begin{equation*}
	\begin{split}
	   \max \sum_{i=0}^{k} (V(q_{i}) - R(q_{i})),
	\end{split}
	\end{equation*}
\end{problem}

subject to incentive compatibility.

\begin{definition}[Incentive Compatibility]
    \textit{The incentive mechanism is incentive compatibility if it is a dominant strategy for each client $i$ and they cannot increase their payoff by misreporting private information regardless of what others do.}
	\begin{equation}
	\label{9}
	U_{client}(q_{i}) \geq U(q_{i} (\hat{\theta_{i}}), \hat{\theta_{i}}).
	\end{equation}
\end{definition}

The aggregation server needs to ensure that targeted clients obtain non-negative payoff in equation (\ref{ir}), \ie satisfy Individual Rationality (IR) constraints as below:

\begin{definition}[Individual Rationality]
	\textit{The incentive mechanism is individually rational if each targeted client $i$ receives a non-negative payoff	by accepting the expected reward $R(q_{i})$ intended for his type, i.e.,}
	\begin{equation}
	\label{ir}
	U_{client}(q_{i}) \geq 0, i \in N.
	\end{equation}
\end{definition}

for every $\theta_{i}$, where $\hat{\theta_{i}} \neq \theta_{i}$. 

\subsection{Optimal Solution}
To solve the problem~\ref{problem:MaxServerComplete}, we differentiate equation (\ref{complete:serverUtility}) with respect to $q_{i}$ and obtain:

\begin{equation}
    \frac{\partial V(q_{i}^{CI})}{\partial q_{i}} - \frac{\partial C_{i}(q_{i}^{CI}, \theta_{i})}{\partial q_{i}} = 0,
\end{equation}

where $q_{i}^{CI}$ denotes the optimal output under complete information (CI). Rearranging this first-order condition yields:

\begin{equation}
    \underbrace{\frac{\partial V(q_{i}^{CI})}{\partial q_{i}}}_{MB_{i}} = \underbrace{\frac{\partial C_{i}(q_{i}^{CI}, \theta_{i})}{\partial q_{i}} }_{MC_{i}}
\end{equation}

Above result suggests that the server increases procuring the output until
its marginal benefit $MB_{i}$ coincides with associated marginal cost $MC_{i}$. 
Since return function $V(q_{i})$ is increasing and concave, its derivative lies
in the positive quadrant but decreases in $q_{i}$. The crossing point between the
marginal benefit and cost functions entails $MB_{i}$ = $MC_{i}$, yielding a socially
optimal output. If this computing contract produces a larger marginal benefit, the $MB_{i}$ function shifts upward, increasing the socially optimal output $q_{i}^{CI}$. In contrast, an increase in the marginal cost of computing, $\frac{\partial C_{i}(q_{i}, \theta_{i})}{\partial q_{i}}$ yields an upward shift in the $MC_{i}$ function, ultimately reducing the socially optimal output $Q_{i}^{CI}$ that the server implements. The optimal output solves $MB_{i}=MC_{i}$, which in this parametric setting entails:

\begin{equation}
    q_{i}^{CI} = \frac{2q_{i}}{(1+\delta \cdot \theta)}
\end{equation}

Solving for output $q_{i}$, we obtain the optimal output:

\begin{equation}
    q_{i}^{CI} = \frac{\lambda (1+\delta \cdot \theta)}{2}
    \label{complete_q}
\end{equation}

\begin{equation}
    R_{i}^{CI} = \frac{1}{1+\delta \cdot\theta} [\frac{\lambda(1+\delta \cdot \theta_{i})}{2}]^{2}
    \label{complete_r}
\end{equation}

\begin{algorithm}[t]
    \caption{AFL in Buyers' Market}
    \label{alg:algorithm}
    \textbf{Input}: FL task $\tau$, historical reputation $\zeta^{\tau-1}$ \\
    \textbf{Output}: Global model $\mathcal{M}_{global}$ 
    \begin{algorithmic}[1] %[1] enables line numbers
        \IF{$\theta_{i}$ is observable}
            \STATE ($q_{i}^{CI}$, $R_{i}^{CI}$) $\rightarrow$ clients $i$ 
        \ELSE
            \STATE ($q_{i}^{*}$, $R_{i}^{*}$) $\rightarrow$ clients $i$ 
        \ENDIF

        \FOR{each clients}
            \IF{$U_{client}(q_{i}) \geq 0$} 
                \STATE $S \leftarrow$ client $i$ \quad //Accept to participate in
            \ENDIF
        \ENDFOR
        \STATE Server determines participants $S = \{ s_{1}, s_{2}, ..., s_{k} \}$
        \STATE Model training $\mathcal{M}_{globel} \leftarrow Aggregation(\mathcal{M}_{local,i})$
        \FOR{each clients}
            \STATE Update reputation $\zeta_{i}^{(\tau)}$
        \ENDFOR
    
        % \STATE Update $\textbf{Re}_{history}$ according to Equation (\ref{contribution_measurement}),(\ref{reputation_calculation_1}),(\ref{reputation_calculation_2})
        \STATE \textbf{return} $\mathcal{M}_{global}$
    \end{algorithmic}
\end{algorithm}

\section{The approach under the Incomplete Information}
\label{sec:incomplete}

\subsection{Problem Formulation}
Consider the aforementioned auction (sec.~\ref{sec:complete}), but now assume that 
the efficiency $\theta_{i}$ of every client $i$ for implementing the project is private information. And we assume that efficiency $\theta_{i} \in [0, 1]$ follows the uniform distribution, which is common knowledge among all players. The aggregation server chooses the output-transfer pair, $(q_{i}, R_{i})$, for each client $i$ to maximize its utility function.

\begin{problem}[Maximize server's utility function under the incomplete information]
\label{problem:MaxServerInComplete}
	\begin{equation*}
	\begin{split}
	   \max \sum_{i=0}^{M} E_{\theta_{i}}(V(q_{i}) - R(q_{i})).
	\end{split}
	\end{equation*}
\end{problem}

\subsection{Optimal Solution}
After some algebra manipulation, the first-order condition with respect
to $q_{i}$ becomes:

\begin{equation}
    \underbrace{\frac{\partial V(q_{i}^{CI})}{\partial q_{i}}}_{MB_{i}} = \underbrace{\frac{\partial C_{i}(q_{i}^{CI}, \theta_{i})}{\partial q_{i}} - \overbrace{(1-\theta_{i}) \frac{\partial^{2} C_{i}(q_{i}^{*}, \theta_{i})}{\partial q_{i} \partial \theta_{i}}}^{Information \thinspace Rent}}_{MC_{i}}
\end{equation}

Since $\frac{\partial V(q_{i}^{*})}{\partial q_{i}}=1$ and given by $\frac{\partial C_{i}}{\partial q_{i}} = \frac{2 q_{i}}{1+2\theta_{i}}$, the optimal output solves $MB_{i}=MVC_{i}$, which in current setting entails:

\begin{equation}
    1 = \frac{2q_{i}}{1+2\theta_{i}}+\left(1-\theta_{i}\right) + \frac{4q_{i}}{\left( 1+2\theta_{i} \right)^{2}}
\end{equation}

Thus, we can obtain the optimal output $q_{i}^{*}$

\begin{equation}
    q_{i}^{*} = \frac{(1+2\theta)^{2}}{6}
\end{equation}

In words, the aggregation server increases procurement until the point at which its
marginal benefit coincides with its associated marginal virtual cost (MVC).
This MVC embodies not only client $i$'s marginal cost but also the information rent that the server needs to provide in order to induce client $i$ report
his type truthfully. From above maximization problem, we can evaluate the transfer of the client $i$ at the optimal output $q_{i}^{*}$, to obtain the optimal transfer to the client $i$ as follows:

\begin{equation}
    R_{i}(q_{i}^{*}) = C_{i}(q_{i}^{*}, \theta_{i}) - (1-\theta_{i})\frac{\partial C_{i}(q_{i}^{*}, \theta_{i})}{\partial \theta_{i}}
\end{equation}

Under a complete information setting, the last term in $MVC_{i}$ (information
rent) was absent. Since the corss-partial derivative $\frac{\partial^{2}C_{i}(q_{i}, \theta_{i})}{\partial q_{i} \partial \theta_{i}}$ is negative, we obtain that $MVC_{i} \leq MC_{i}$. Therefore, the socially optimal output under complete information is larger than that under incomplete, $q_{i}^{CI} \leq q_{i}^{*}$. Intuitively, the server must pay an information rent to all bidders to induce truthful revelation of their types, incurring more costs to implement the auction than under complete
information, ultimately inducing lower output levels. This is commonly referred
in the literature as downward distortion for all bidders with efficiency levels $\theta_{i} \neq 1$.

However, the output of the bidder with the highest efficiency, $\theta_{i}=1$, suffers no distortion when moving from a complete to an incomplete information context. Indeed, $MVC_{i}$ simplifies to $MC_{i}$ when evaluated at $\theta_{i}=1$, so the first-order conditions across information contexts coincide, and $q_{i}^{CI} = q_{i}^{*}$. Intuitively, the most efficient bidder has no incentives to underreport his valuation at $\theta_{i}=1$. This result is known as \textit{no distortion at the top}.

\section{Reputation Mechanism Design}
In the client selection phase, we put forward a reputation mechanism for the aggregation server to choose clients with high reliability and data quality, while also reducing vulnerabilities to external risks such as poisoning attacks. To ensure the process is trustworthy, we incorporate blockchain technology to permanently and transparently log each client's reputation scores over time. Our reputation mechanism consists of two principal components: initial contribution measurement to assess each client's performance; followed by reputation calculation to derive scores based on measured contributions. By integrating blockchain in this manner, the selection process runs with full visibility and prevents any distortion of reputations for any purpose. 

% In the decision phrase, we propose a reputation mechanism for the aggregation server to select clients with high reliability and data quality, and also to prevent from external attack (\eg poisoning attack). To further make the reputation mechanism more trustworthy, we leverage blockchain to record and make reputation scores of clients immutable. Our reputation mechanism can be divided into two steps: contribution measurement and reputation calculation.

% Reputation score reflects how efficient a candidate client is and the data quality it provides, which is an important metric for the aggregation server to select clients. Intuitively, a client's contribution to FL task directly indicates its quality, and we consider it as an important evaluation metric for the client's reputation. Our reputation mechanism is divided into two steps: contribution measurement and reputation calculation.
\subsection{Contribution Measurement}
As a fairness valuation method, banzhaf index~\cite{banzhaf1964weighted} from cooperative game theory can measure individual influence in collective decision making. As a result, we leverage banzhaf index as an efficient way to measure the contribution of each client in FL and formulate it as follows:

\begin{equation}
\zeta_{i}^{(\tau)}=\frac{1}{2^{n-1}} \sum_{S \subseteq N \backslash i}[U_{server}(S \cup i)-U_{server}(S)],
    \label{contribution_measurement}
\end{equation}

where $\zeta_{i}^{(\tau)}$ represents the contribution of client $i$ in the FL task $\tau$.
% where $N=\{s_{1}, s_{2}, ..., s_{n}\}$.

% Banzhaf method\cite{} is an effective approach for contribution assessment that achieves stable performance\cite{} in cooperative game theory. We employ the Banzhaf method to evaluate client's contribution to FL tasks.

% $$(C_{1}^{(\tau)},...,C_{n}^{(\tau)})=Banzhaf(w_1,...,w_n,\mathcal{M}_{global})$$
% where $C_{i}^{(\tau)}$ represents the contribution of $i$ in the $\tau$ task.

\subsection{Reputation Calculation}
To effectively evaluate clients' reputation, we normalize the contributions $\zeta_{i}^{(\tau)}$ of clients $i$ to reputation scores $\varepsilon_{i}^{\tau}$. Inspired by~\cite{zhang2021pricing}, it is intuitive to give higher weights to more recent reputation records. The reputation score is calculated by equation (\ref{equation:reputation}).

\begin{equation}
\varepsilon_{i}^{(\tau)} = \varepsilon_{i}^{(\tau-1)}*w_1 + \zeta_{i}^{(\tau)}*w_2,
\label{equation:reputation}
\end{equation}

where $w_1$ and $w_2$ represent the reputation weights.

\begin{table*}[h]
\centering
\caption{}
\begin{tabular}{ccccccc}
\hline
\multicolumn{7}{c}{\textbf{MNIST}} \\
\hline

 & $k$=10  &$k$=15   &$k$=20 & $k$=25 & $k$=30 & $k$=35  \\ \hline
Our Approach (Complete)  & 9,367.1 $\pm$ 8.9 & 14,039.8 $\pm$ 12.2  & 18,699.2 $\pm$ 19.6  & 23,358.5 $\pm$ 21.0 & 27,983.2 $\pm$ 27.7 & 32,618.3 $\pm$ 33.5\\
Our Approach (Incomplete)& 9,321.9 $\pm$ 10.9 & 13,986.5 $\pm$ 8.9  & 18,624.6 $\pm$ 18.4  & 23,269.9 $\pm$ 25.0 & 27,891.7 $\pm$ 32.7 & 32,515.4 $\pm$ 33.3\\
Randomized Auction  & 9,021.8 $\pm$ 31.0 & 13,240.1 $\pm$ 334.1 & 17,602.6 $\pm$ 557.9 & 22,443.2 $\pm$ 70.6 & 26,838.1 $\pm$ 54.0 & 31,154.5 $\pm$ 110.8\\
Price First  & 8,949.2 $\pm$ 12.3 & 13,399.9 $\pm$ 7.9  & 17,822.2 $\pm$ 27.9  & 22,277.3 $\pm$ 32.5 & 26,673.7 $\pm$ 39.0 & 31,077.3 $\pm$ 38.8\\ \hline

\hline
\multicolumn{7}{c}{\textbf{Fashion MNIST}} \\ \hline
 & $k$=10  &$k$=15   &$k$=20 & $k$=25 & $k$=30 & $k$=35  \\ \hline
Our Approach (Complete)  & 8,575.1 $\pm$ 24.3 & 12,798.8 $\pm$ 39.3   & 16,996.9 $\pm$ 18.2  & 21,123.1 $\pm$ 30.5  & 25,284.9 $\pm$ 51.5  & 29,355.9 $\pm$ 78.9\\
Our Approach (Incomplete)& 8,560.2 $\pm$ 19.7 & 12,785.1 $\pm$ 22.3   & 16,946.3 $\pm$ 18.9  & 21,108.5 $\pm$ 47.3  & 25,269.3 $\pm$ 48.0  & 29,352.3 $\pm$ 61.1\\
Randomized Auction        & 8,052.1 $\pm$ 301.0 & 12,040.9 $\pm$ 280.5 & 16,150.3 $\pm$ 106.2 & 20,188.5 $\pm$ 116.5 & 24,184.9 $\pm$ 66.6 & 27,824.3 $\pm$ 675.5\\
Price First       & 8,230.5 $\pm$ 29.5 & 12,274.5 $\pm$ 56.1   & 16,288.4 $\pm$ 61.1 & 20,308.0 $\pm$ 43.6  & 24,231.8 $\pm$ 61.8  & 28,143.8 $\pm$ 154.4\\ \hline

\hline
\multicolumn{7}{c}{\textbf{CIFAR-10}} \\ \hline
 & $k$=10  &$k$=15   &$k$=20 & $k$=25 & $k$=30 & $k$=35  \\ \hline
Our Approach (Complete)  & 6,135.0 $\pm$ 98.3  & 8,434.1 $\pm$ 191.1 & 11,496.5 $\pm$ 76.6 & 14,121.0 $\pm$ 131.9 & 17,616.8 $\pm$ 329.5 & 20,591.3 $\pm$ 277.7\\
Our Approach (Incomplete) & 6,040.3 $\pm$ 135.0  & 8,309.1 $\pm$ 91.6 & 11,301.0 $\pm$ 71.7 & 14,109.7 $\pm$ 287.7 & 17,542.7 $\pm$ 116.2 & 20,573.0 $\pm$ 199.7\\
Randomized Auction          & 5,341.3 $\pm$ 421.3  & 7,356.5 $\pm$ 227.8 & 10,503.7 $\pm$ 301.1 & 12,968.1 $\pm$ 325.0 & 15,790.5 $\pm$ 260.5 & 18,482.6 $\pm$ 200.3\\
Price First        & 4,937.7 $\pm$ 141.5  & 6,787.8 $\pm$ 70.8 & 9,160.9 $\pm$ 175.7  & 11,418.7 $\pm$ 84.7 & 14,161.2 $\pm$ 119.1 & 16,396.7 $\pm$ 237.8\\ \hline

\end{tabular}
\label{Q1}
\end{table*}

\section{Experiments}

To validate the efficiency of our approach, we aim to answer the following questions in this section.

\begin{itemize}
    \item \textbf{Q1: Performance Improvement.} Whether our approach has better performance compared to baseline methods?
    \item \textbf{Q2: Poisoning Attack Detection.} Can our approach prevent from poisoning attack? 
    \item \textbf{Q3: Universality.} Does our approach work well with different aggregation algorithms?
    \item \textbf{Q4: Robustness.} Is our approach robust enough to protect from external attack?
\end{itemize}

These questions are examined in experiments on MNIST~\cite{lecun1998gradient}, Fashion MNIST~\cite{xiao2017fashion} and CIFAR-10~\cite{krizhevsky2009learning}. 
We have established a total of $m$ clients in the federated learning (FL) ecosystem, and datasets are divided among these $m$ clients. The model trained on MNIST consists of three fully connected layers. For the model trained on Fashion MNIST, it consists of two convolutional layers and two fully connected layers. For CIFAR-10, we use the exact architecture of MobileNet\cite{howard2017mobilenets} in their open sourced code.

% The input layer has 784 neurons, each hidden layer has 512 neurons and the output layer has 10 neurons.  The input layer has 800 neurons, each hidden layer has 500 neurons and the output layer has 10 neurons. 

% The MNIST dataset consists of 60,000 grayscale images of handwritten digits from 0 to 9. Fashion-MNIST is a dataset containing 60,000 grayscale images of fashion items, such as clothing, footwear, and accessories. CIFAR-10 is a dataset consisting of 60,000 color images across 10 different classes, including airplanes, cars, etc. 

\begin{figure}
    \setlength{\abovecaptionskip}{-0.10cm}
	\centering{\includegraphics[scale=0.3]{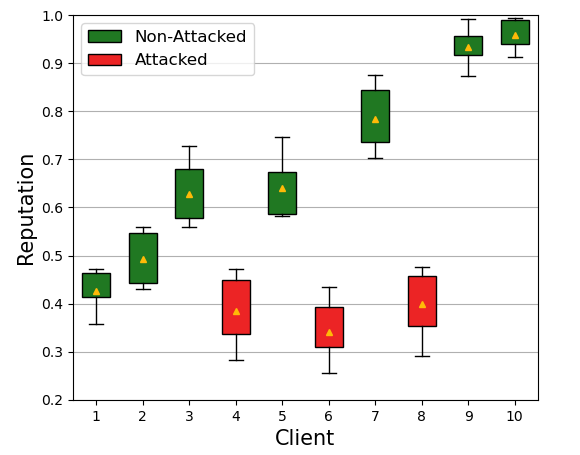}}
    \caption{Reputation of non-attacked clients and poisoning attacked clients.}
	\label{poisoningAttack}
\end{figure}

\begin{figure*}
\setlength{\abovecaptionskip}{-0.10cm}
  \centering
  \begin{subfigure}
  \centering{\includegraphics[scale=0.28]{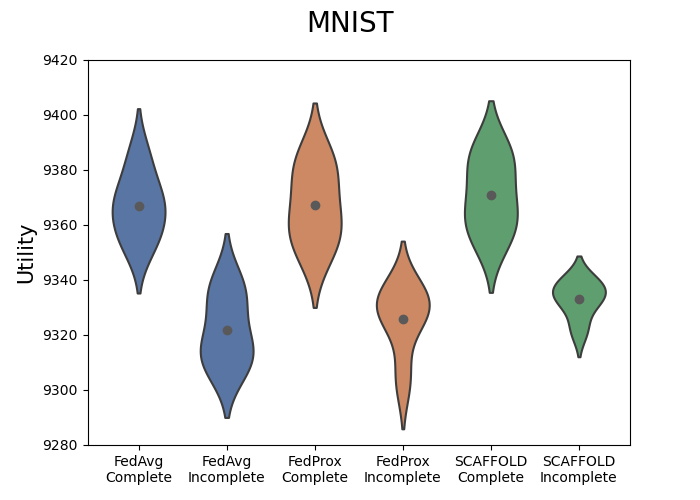}}
  \label{fig:subfig1}
  \end{subfigure}
  \begin{subfigure}
  \centering{\includegraphics[scale=0.28]{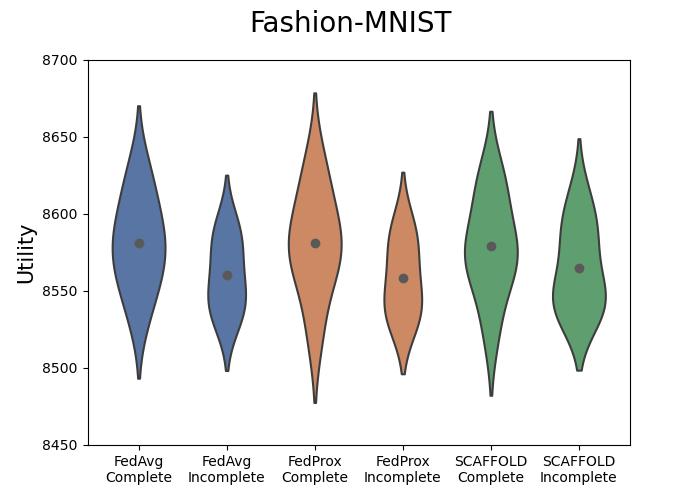}}
  % \caption{xx}
  \label{fig:subfig1}
  \end{subfigure}
  \begin{subfigure}
  \centering{\includegraphics[scale=0.28]{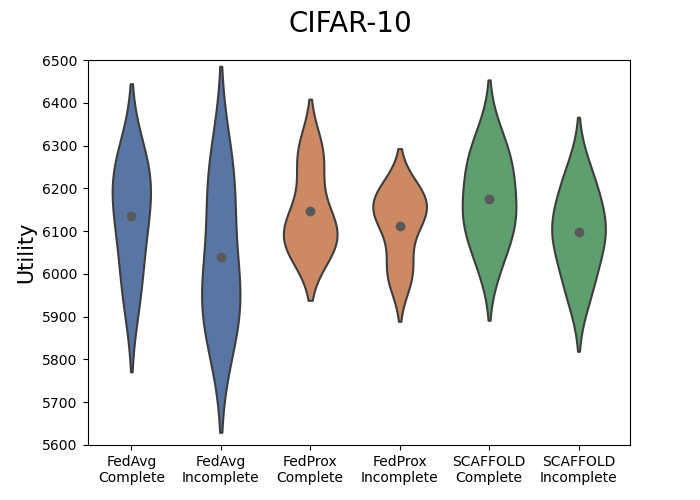}}
  % \caption{xx}
  \label{aggregationAlgorithm}
  \end{subfigure}
  
  \caption{The utility of the aggregation server with different aggregation algorithms.}
  \label{avgAlgorithm}
\end{figure*}

\begin{figure*}
\setlength{\abovecaptionskip}{-0.10cm}
  \centering
  \begin{subfigure}
  \centering{\includegraphics[scale=0.18]{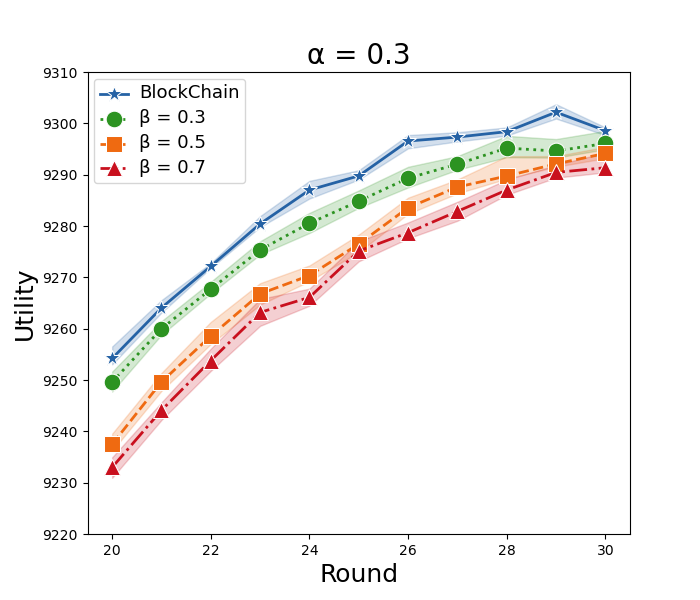}}
  \label{fig:subfig1}
  \end{subfigure}
  \begin{subfigure}
  \centering{\includegraphics[scale=0.18]{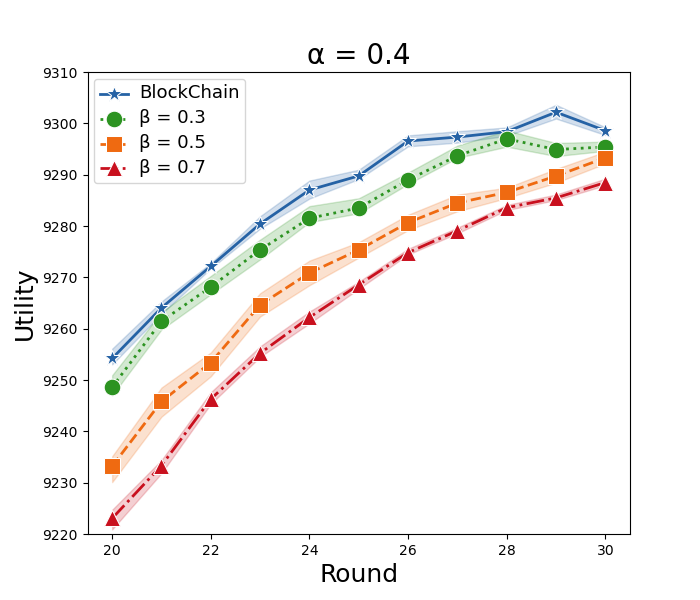}}
  % \caption{xx}
  \label{fig:subfig1}
  \end{subfigure}
  \begin{subfigure}
  \centering{\includegraphics[scale=0.18]{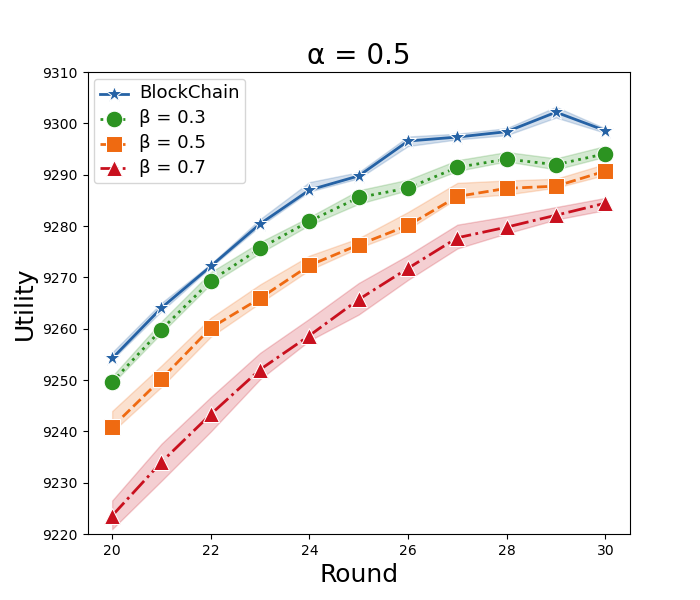}}
  % \caption{xx}
  \label{fig:subfig1}
  \end{subfigure}
  \begin{subfigure}
  \centering{\includegraphics[scale=0.18]{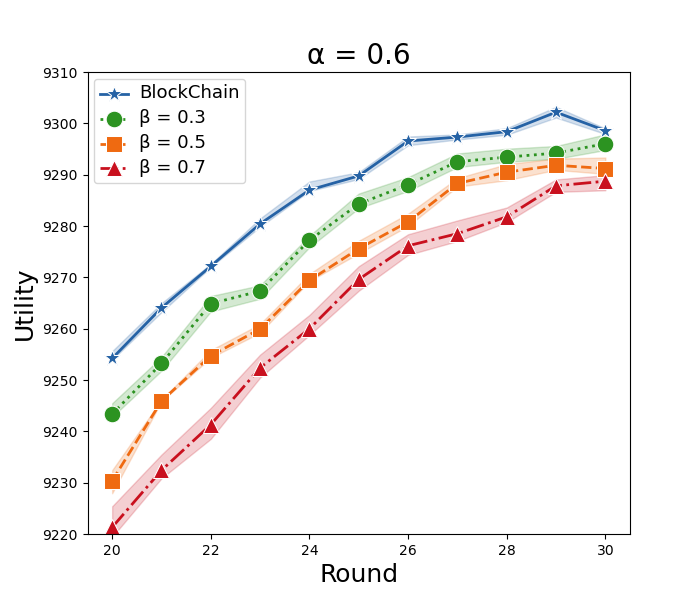}}
  % \caption{xx}
  \label{fig:subfig1}
  \end{subfigure}
  \begin{subfigure}
  \centering{\includegraphics[scale=0.18]{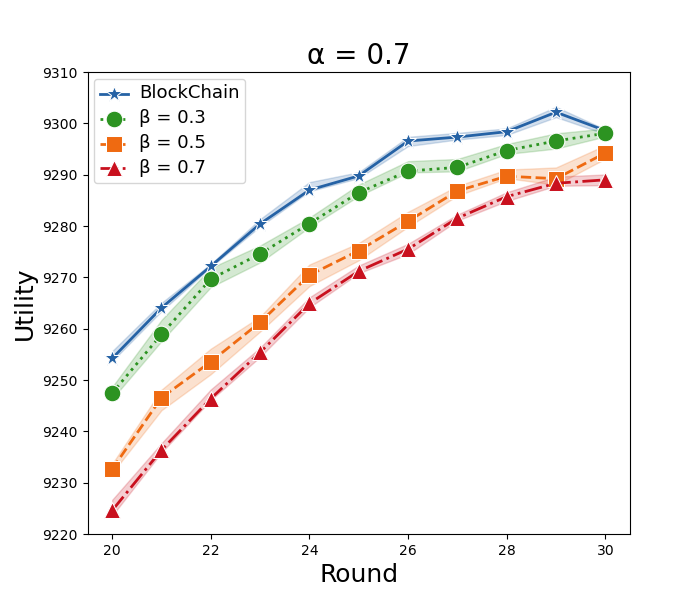}}
  % \caption{xx}
  \label{fig:subfig1}
  \end{subfigure}
  \begin{subfigure}
  \centering{\includegraphics[scale=0.18]{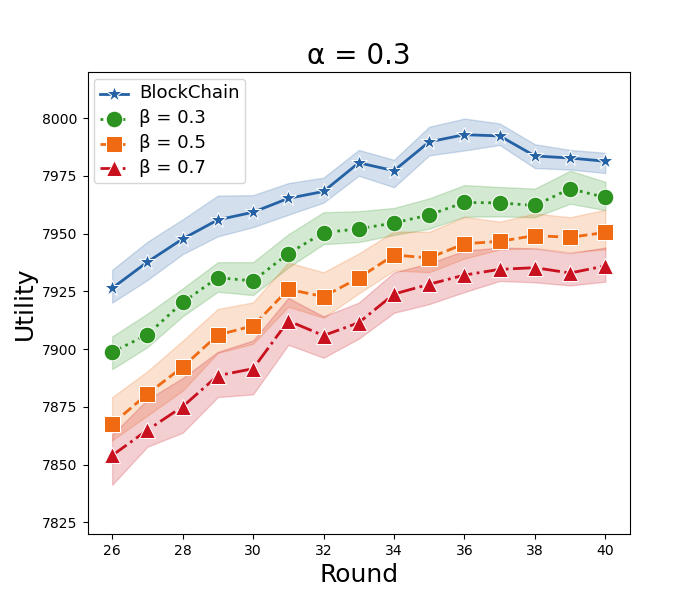}}
  % \caption{xx}
  \label{fig:subfig1}
  \end{subfigure}
  \begin{subfigure}
  \centering{\includegraphics[scale=0.18]{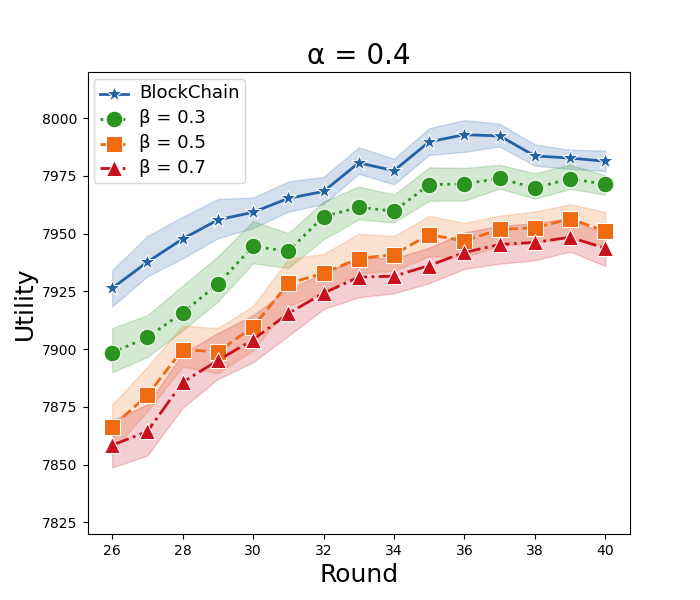}}
  % \caption{xx}
  \label{fig:subfig1}
  \end{subfigure}
  \begin{subfigure}
  \centering{\includegraphics[scale=0.18]{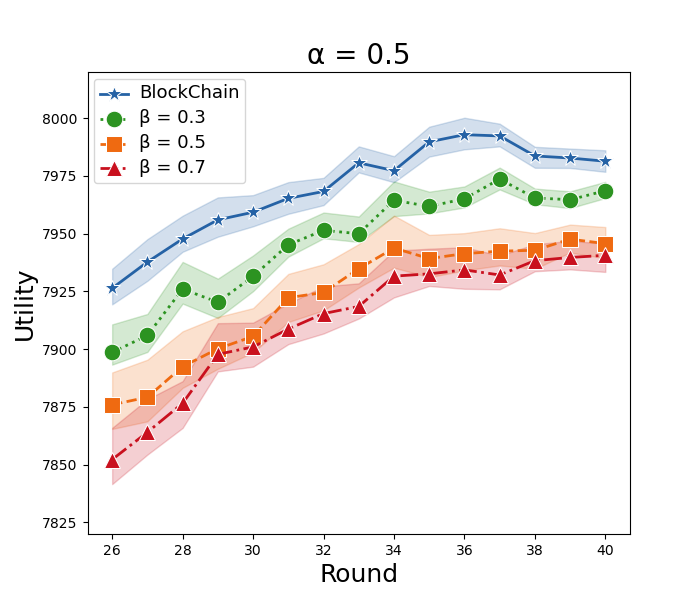}}
  % \caption{xx}
  \label{fig:subfig1}
  \end{subfigure}
  \begin{subfigure}
  \centering{\includegraphics[scale=0.18]{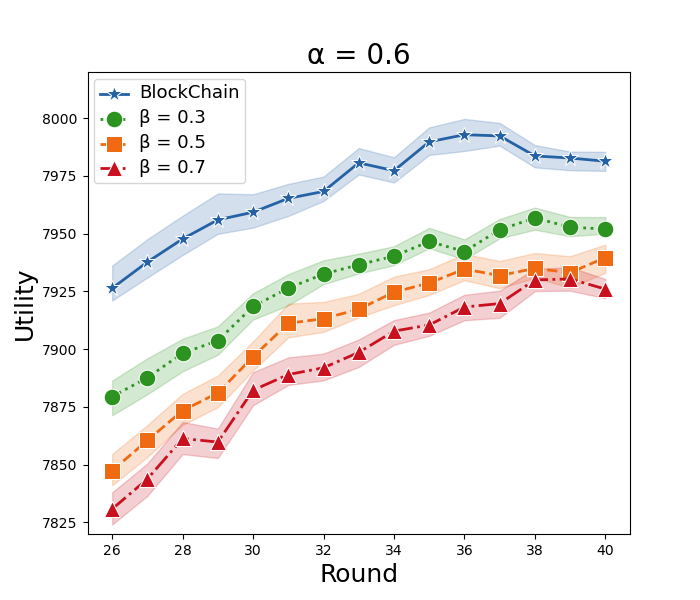}}
  % \caption{xx}
  \label{fig:subfig1}
  \end{subfigure}
  \begin{subfigure}
  \centering{\includegraphics[scale=0.18]{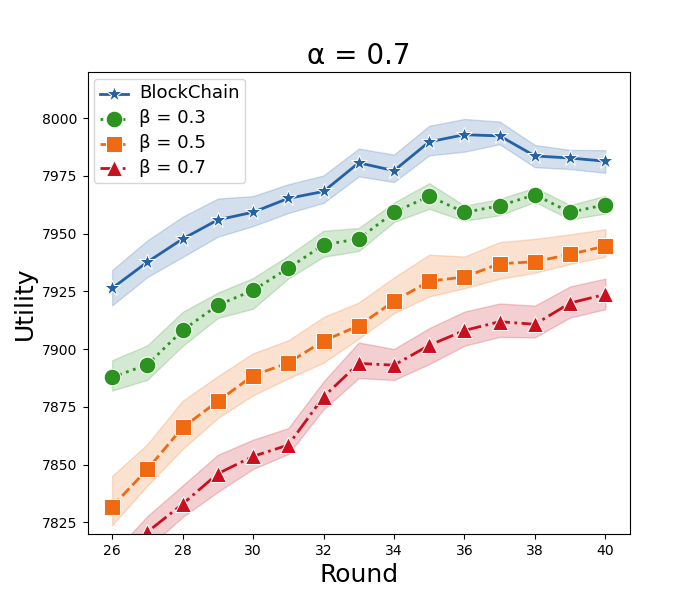}}
  % \caption{xx}
  \label{fig:subfig1}
  \end{subfigure}
  \begin{subfigure}
  \centering{\includegraphics[scale=0.18]{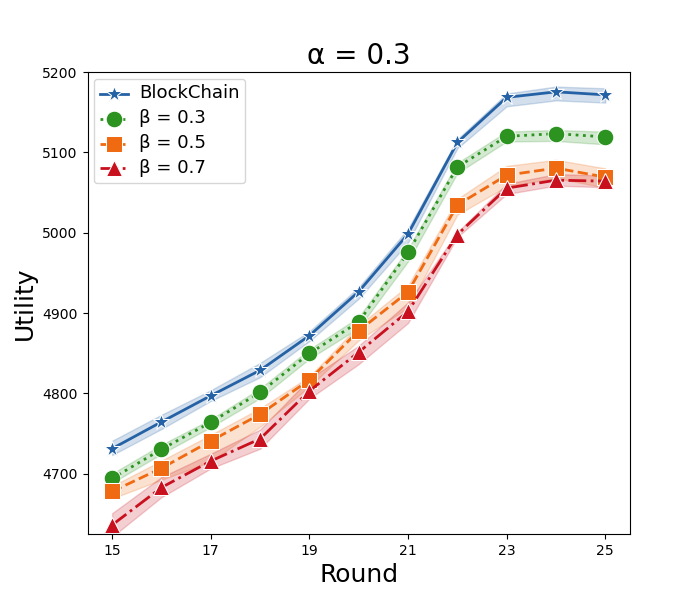}}
  % \caption{xx}
  \label{fig:subfig1}
  \end{subfigure}
  \begin{subfigure}
  \centering{\includegraphics[scale=0.18]{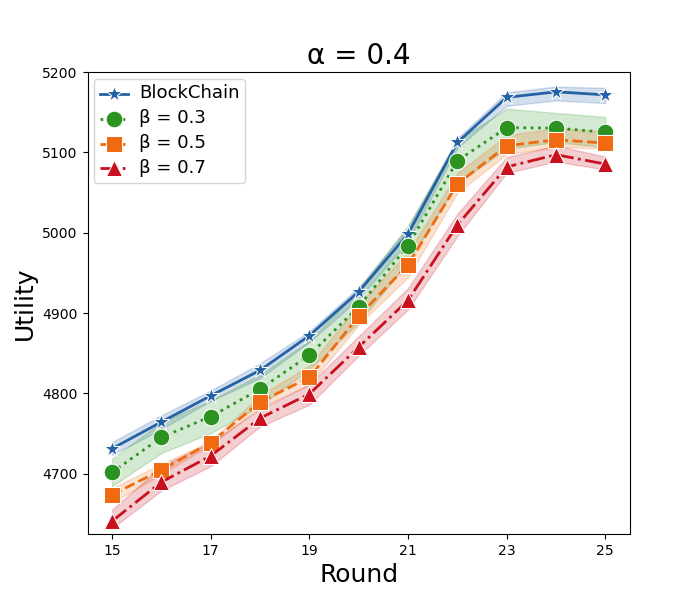}}
  % \caption{xx}
  \label{fig:subfig1}
  \end{subfigure}
  \begin{subfigure}
  \centering{\includegraphics[scale=0.18]{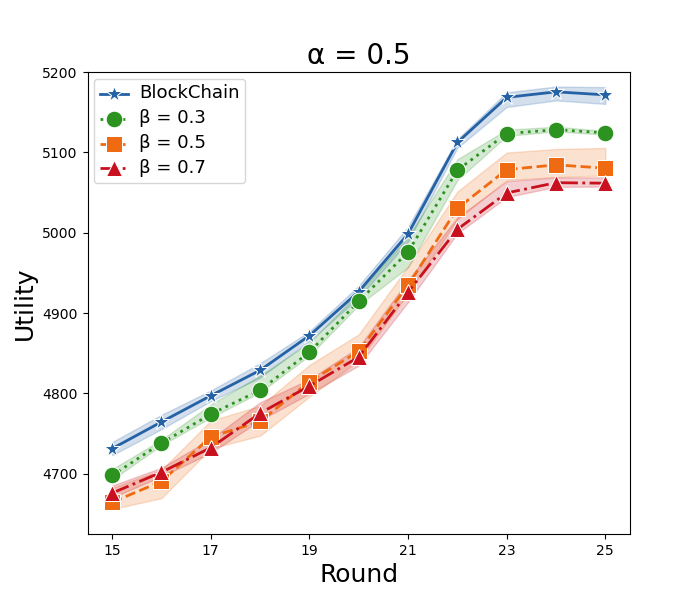}}
  % \caption{xx}
  \label{fig:subfig1}
  \end{subfigure}
  \begin{subfigure}
  \centering{\includegraphics[scale=0.18]{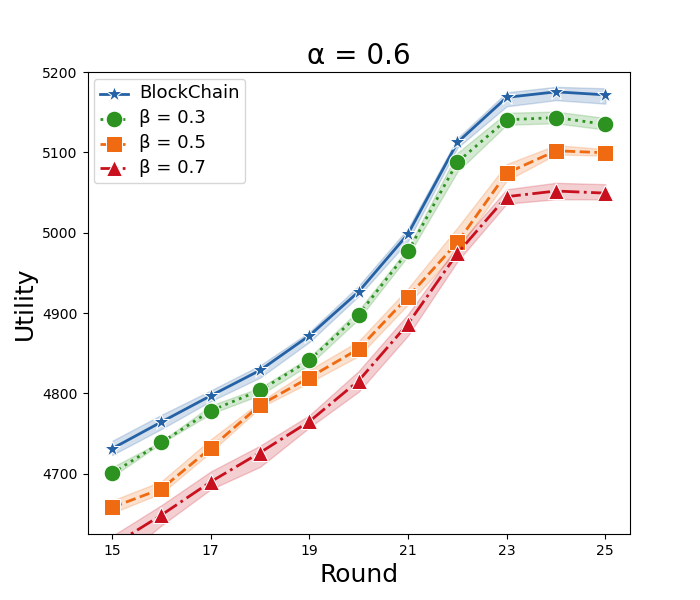}}
  % \caption{xx}
  \label{fig:subfig1}
  \end{subfigure}
  \begin{subfigure}
  \centering{\includegraphics[scale=0.18]{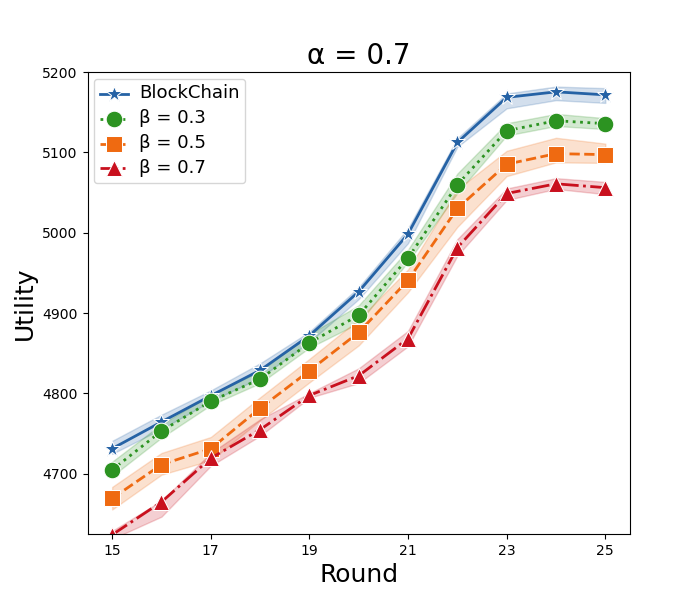}}
  % \caption{xx}
  \label{fig:subfig1}
  \end{subfigure}
  
  \caption{The utility of the aggregation server with blockchain-based reputation mechanism and vulnerable reputation mechanism. $\alpha$ is the proportion of attacked clients out of the total number of clients, and $\beta$ is the degree of attack.}
  \label{blockchain}
\end{figure*}

\subsection{Performance Improvement (Q1)}
To evaluate the utility that the aggregation server obtain and compare with baseline methods: price first~\cite{le2021incentive} and randomized auction~\cite{li2017crowdsourcing}, we conduct experiments with different number of clients $k$ selected clients in FL ecosystem under the various datasets and leverage FedAvg for aggregation. The results shown in TABLE~\ref{Q1} demonstrate that our approach outperforms the baseline methods.
Given the presence of information asymmetry, we consider the performance achieved under the complete information scenario as the ground-truth. Our approach under the incomplete information scenario demonstrates a close resemblance to this ground-truth performance. It also indicates that paying information rent significantly reduces the disparity between the utility under incomplete information and complete information.

\subsection{Poisoning Attack Detection (Q2)}
To verify our approach is trustworthy in preventing from poisoning attackers, we simulate three poisoning attacked clients in FL ecosystem. We calculate their reputation value by our approach and observe in Fig.~\ref{poisoningAttack} that the reputation of the poisoning attacked clients is lower than others'.

\subsection{Universality (Q3)}
Considering different FL settings, our approach needs to perform well across different aggregation algorithms.
To evaluate its effectiveness, we examine our approach under three aggregation algorithms: FedAvg, FedProx~\cite{li2020federated}, and Scaffold~\cite{karimireddy2020scaffold}, with both complete and incomplete information.
The results in Fig.~\ref{avgAlgorithm} demonstrate that our approach works well and has similar performance in different FL settings. 

\subsection{Robustness (Q4)}
As any self-interest client may have incentive to cheat reputation, the robustness of the blockchain-based reputation mechanism needs to be examined. We conduct experiments with various degrees of attacks and different ratios of attacked clients on reputation records.
By comparing the reputation mechanisms without recording reputation on blockchain, the results in Fig.~\ref{blockchain} show that by recording the reputation on blockchain, the aggregation server can get higher profit and establish a robust reputation mechanism.

% demonstrate that by leveraging the information rent, our approach under incomplete information can achieve similar performance comparable to that of the complete information scenario.

% By considering different aggregation algorithms in FL, it requires our approach to work well when using different aggregation algorithms. We try to examine our approach under the different aggregation algorithm: FedAvg~\cite{}, FedProx~\cite{} and Scaffold~\cite{} in both complete and incomplete information scenario. Results in Fig.~\ref{aggregationAlgorithm} shows that by paying the information rent our approach under the incomplete information can obtain approximate performance to complete information scenario.

\section{Conclusion}
% From a new perspective, we consider the AFL under buyers' market. The approach adapts a procurement auction framework and utilizes blockchain-based reputation to incentivize high quality clients to participate in. Theoretical analysis and experimental results demonstrate that the approach achieves desirable properties and outperform the baseline methods.

Casting aside established preconceptions, this paper applies an innovative analytical angle to gain new insights into the AFL incentive mechanisms under the market forces of buyers. We adopt procurement auction to approach the scenario where clients compete with one another to win the computing contract, and utilize blockchain-based reputation to select reliable candidates. Through experimental validation, our proposed design is shown to achieve desirable properties and outperform baseline approaches.

% references
\bibliographystyle{ieeetr} 
\bibliography{references.bib}

\end{document}